\documentclass[sigconf]{acmart}

\usepackage{amsthm}
\usepackage{amsmath}
\usepackage{algorithm}
\usepackage{enumitem}
\usepackage{picinpar}
\usepackage{lineno}
\usepackage{graphicx}

\usepackage{algorithmicx}
\usepackage[noend]{algpseudocode}
\usepackage{subfigure}
\usepackage{multirow}
\usepackage{color}
\usepackage{balance}
\usepackage{enumitem}
\usepackage{hhline}
\usepackage[normalem]{ulem}
\usepackage{booktabs}
\usepackage{wrapfig}
\usepackage{cancel}
\usepackage{hyperref}
\usepackage{makecell}

\newcommand{\bL}{\ensuremath{\mathcal{L}}}

\newcommand{\bT}{\ensuremath{\mathcal{T}}}

\renewcommand{\vec}[1]{\ensuremath{\mathbf{#1}}}

\newcommand{\stitle}[1]{\vspace{1mm} \noindent {\bf #1}}

\newcommand{\eg}{{\it e.g.}}

\newcommand{\ie}{{\it i.e.}}

\newcommand{\method}[1]{\textsc{#1}}
\newcommand{\model}{\method{GCoT}{}}

\newcommand{\eat}[1]{}

\newcommand{\stkout}[1]{\ifmmode\text{\sout{\ensuremath{#1}}}\else\sout{#1}\fi}

\AtBeginDocument{%
  }

\copyrightyear{2025}
\acmYear{2025}
\setcopyright{acmlicensed}
\acmConference[KDD '25] {Proceedings of the 31st ACM SIGKDD Conference on Knowledge Discovery and Data Mining V.1}{August 3--7, 2025}{Toronto, ON, Canada.}
\acmBooktitle{Proceedings of the 31st ACM SIGKDD Conference on Knowledge Discovery and Data Mining V.1 (KDD '25), August 3--7, 2025, Toronto, ON, Canada}
\acmISBN{979-8-4007-1245-6/25/08}
\acmDOI{10.1145/3711896.3736974}

\renewcommand{\shortauthors}{Trovato et al.}

\begin{document}

\title{GCoT: Chain-of-Thought Prompt Learning for Graphs}

\author{Xingtong Yu}
\affiliation{%
 \institution{Singapore Management University}
  \country{Singapore}}
\email{xingtongyu@smu.edu.sg}

\author{Chang Zhou}
\affiliation{%
 \institution{University of Science and Technology of China}
 \country{China}}
\email{zhouchang21sy@mail.ustc.edu.cn}

\author{Zhongwei Kuai}
\affiliation{%
 \institution{University of Science and Technology of China}
 \country{China}}
\email{asagiri@mail.ustc.edu.cn}

\author{Xinming Zhang}
\affiliation{%
  \institution{University of Science and Technology of China}
  \country{China}}
\email{xinming@ustc.edu.cn}

\author{Yuan Fang}
\affiliation{%
  \institution{Singapore Management University}
  \country{Singapore}}
\email{yfang@smu.edu.sg}



\begin{abstract}
Chain-of-thought (CoT) prompting has achieved remarkable success in natural language processing (NLP). However, its vast potential remains largely unexplored for graphs. This raises an interesting question: How can we design CoT prompting for graphs to guide graph models to learn step by step?
On one hand, unlike natural languages, graphs are non-linear and characterized by complex topological structures. On the other hand, many graphs lack textual data, making it difficult to formulate language-based CoT prompting. 
In this work, we propose the first CoT prompt learning framework for text-free graphs, \model. Specifically, we decompose the adaptation process for each downstream task into a series of inference steps, with each step consisting of prompt-based inference, ``thought'' generation, and thought-conditioned prompt learning. While the steps mimic CoT prompting in NLP, the exact mechanism differs significantly. Specifically, at each step, an input graph, along with a prompt, is first fed into a pre-trained graph encoder for prompt-based inference. We then aggregate the hidden layers of the encoder to construct a ``thought'', which captures the working state of each node in the current step.  Conditioned on this thought, we learn a prompt specific to each node based on the current state. These prompts are fed into the next inference step, repeating the cycle.
To evaluate and analyze the effectiveness of \model, we conduct comprehensive experiments on eight public datasets, which demonstrate the advantage of our approach. 
\end{abstract}





\maketitle

\newcommand\kddavailabilityurl{https://doi.org/10.5281/zenodo.15501903}

\ifdefempty{\kddavailabilityurl}{}{
\begingroup\small\noindent\raggedright\textbf{KDD Availability Link:}\\
The source code of this paper has been made publicly available at \url{\kddavailabilityurl}.
\endgroup
}

\section{Introduction}
\begin{figure}[t]
\centering
\includegraphics[width=1\linewidth]{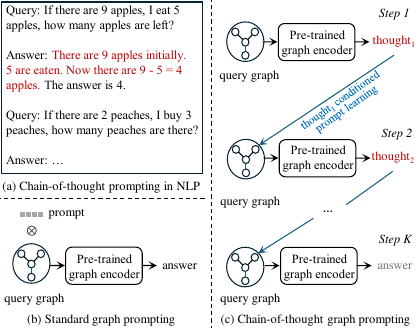}
\caption{Illustration of chain-of-thought in NLP, standard graph prompts and graph chain-of-thought prompting.}
\label{fig.intro-motivation}
\end{figure}

Recently, Chain-of-Thought (CoT) prompting has demonstrated significant advancement in the field of natural language processing (NLP) \cite{wei2022chain,wang2023self,chu2023survey}, which mimics the logical process a human may employ to solve a task. Instead of directly providing an answer, CoT prompting decomposes a problem into several steps, guiding the pre-trained language model to follow these steps that lead to the final answer. For example, 
given the math question in Fig.~\ref{fig.intro-motivation}(a), a language model utilizes CoT prompting (\eg, initial quantity, eaten quantity, and subtraction) to reach the final answer of 4.

However, the vast potential of CoT prompt learning remains unexplored on pre-trained graph models. Graphs can capture the interactions between entities in a wide range of domains, exhibiting a non-linear topological structure, such as molecular graphs \cite{wang2023automated,lee2023shift,li2024finemoltex}, citation networks \cite{kanakia2019scalable,xiong2017explicit,DBLP:conf/aaai/LiWLS24}, and social networks \cite{ji2023community,zhang2023capacity}.
Conventional approaches typically retrain graph neural networks (GNNs) \cite{kipf2016semi,velivckovic2017graph} or graph transformers \cite{ying2021transformers,yun2019graph} for each specific task in an end-to-end supervised manner, relying on abundant labeled data.
More recent pre-training methods \cite{velickovic2019deep,you2020graph} learn task-invariant, general properties from unlabeled graphs through self-supervised pretext tasks and are then fine-tuned via task-specific labels to adapt to various downstream applications \cite{hu2020gpt,qiu2020gcc}. To further narrow the gap between pre-training and downstream tasks, prompt learning \cite{yu2024few,liu2023graphprompt,fang2024universal} has emerged as a low-resource alternative. They unify the objectives of pre-training and downstream tasks using the same template and employ a lightweight prompt to modify either the input features or hidden embeddings of the pre-trained graph encoder, while keeping the pre-trained weights frozen, as shown in Fig.~\ref{fig.intro-motivation}(b).
A contemporary work \cite{jin2024graph} leverages graphs to guide reasoning; however, it is built on the CoT mechanism from NLP and relies on textual data to construct thoughts, precluding its application to general \emph{text-free} graphs.
For text-free graphs, existing graph learning methods---including supervised, pre-training, and prompting approaches---produce a ``final answer'' in a single inference step, which may limit the refinement of the final prediction. In this work, we explore the following question: \textit{Would introducing additional inference steps in a CoT style enhance the ability of pre-trained graph models to refine their predictions?} 

Due to the significant differences between language and graph data, replicating CoT prompting from NLP for graphs is challenging. In NLP, a CoT prompt can be handcrafted before the learning phase and typically consists of a structured text in the form \(\langle \text{input, chain of thought, output} \rangle\) \cite{wei2022chain}. This prompt serves as an example to guide the model in generating intermediate thoughts that lead to the final answer. In contrast, our work explores a different prompting format for text-free graphs, which mimics the CoT approach in NLP but is not a direct application. Instead of designing prompts prior to the learning phase, \textit{we generate them step by step based on intermediate ``thoughts''}, improving the model's inference ability on downstream tasks by incorporating additional inference steps while freezing the pre-trained weights. To realize this vision, we must address two questions.

First, \textit{what should be the inference steps and thoughts for a graph task?} In NLP, a thought is defined as a short instruction that reflects a single reasoning step, with each intermediate textual answer serving as a thought \cite{chu2023survey}. However, in general text-free graphs, we cannot directly leverage text as prompts or thoughts. 
In this work, we aim to improve the inference capability of a pre-trained graph model by incorporating additional steps to refine the answer. We design an \emph{inference step} with three substages: prompt-based inference, thought construction, and prompt learning. For prompt-based inference, we feed the input graph for the downstream task, along with some prompts, into a pre-trained graph encoder. Then, 
we construct a ``thought'' by fusing embeddings from each layer of the pre-trained encoder
to capture the current working state with varying levels of topological knowledge \cite{kipf2016semi}. Lastly, the thought is used to learn a set of prompts that guide the next step.

Second, \textit{how can we leverage a ``thought'' to learn prompts and guide the next-step inference?} In NLP, CoT prompting is typically implemented by appending specific phrases such as ``let's think step by step'' or by providing few-shot CoT examples \cite{wei2022chain,feng2024towards}. Then, following a given prompt template, the language model generates new thoughts based on the query and prior thoughts, which in turn facilitate the next reasoning step. 
In our work, the absence of textual data prevents us from explicitly guiding the next step. 
Moreover, since each node in a graph exhibits unique characteristics, inference may benefit from node-specific prompts. Thus, we propose a \emph{thought-conditioned prompt learning} method to guide the next inference step. Specifically, inspired by conditional prompt learning \cite{zhou2022conditional}, we generate a unique prompt for each node via a conditional network (condition-net), which is conditioned on the node-specific element of the previously constructed thought. The generated prompts then feed into the graph encoder to initiate and guide the next step, repeating the process. 

In summary, the contributions of this work are fourfold.
(1)  We propose \model, a Graph CoT prompting approach to guide pre-trained graph models to perform step-by-step inference. To the best of our knowledge, this is the first exploration of CoT-style prompting on text-free graphs.
(2) We design an inference step with three substages: prompt-based inference, thought construction, and thought-conditioned prompt learning. In particular, a thought is constructed by fusing embeddings from each layer of the graph encoder to capture fine-grained topological knowledge in each step. 
(3) We employ a condition-net to generate node-specific prompts based on the previous thought, enabling the model to perform inference for the downstream task through a step-by-step, node-specific adaptation.
(4) We conduct extensive experiments on eight benchmark datasets, demonstrating the superior performance of \model\ compared to a suite of state-of-the-art methods.
\section{Related Work}

In this section, we briefly review related work on CoT prompting, graph learning, and graph prompt learning.

\stitle{Chain-of-Thought prompting.}
Chain-of-Thought (CoT) prompting has emerged as a groundbreaking technique in NLP, empowering language models to address complex reasoning tasks by producing intermediate reasoning steps \cite{wei2022chain,wang2023self,gao2023pal}. By breaking down problems into a sequence of logical steps, CoT emulates human-like thought processes, leading to significant improvements in model performance on tasks requiring structured reasoning \cite{yao2024tree,feng2024towards}. Despite its remarkable success in NLP applications, the potential of CoT prompting for graphs remains largely unexplored.  A contemporary work, GraphCoT \cite{jin2024graph}, leverages the inherent relational information in text-attributed graphs \cite{yan2023comprehensive,wen2023augmenting} to guide the reasoning process. However, GraphCoT primarily focuses on question-answering tasks for natural language and cannot be extended to general text-free graphs that lack textual descriptions.

\stitle{Graph representation learning.}
GNNs \cite{kipf2016semi, velivckovic2017graph,yu2023learning,jiang2024ragraph,jiang2023uncertainty} are the dominant technique for graph representations learning. They generally update node embeddings iteratively by aggregating information from their local neighborhoods based on a message-passing mechanism \cite{xu2018powerful, hamilton2017inductive,DBLP:conf/www/LiWXFWLS24}. Despite their success, GNNs often demand substantial amounts of task-specific labeled data and necessitate retraining for each new task, limiting their flexibility and scalability.
Recently, researchers have extensively investigated pre-training techniques for graphs \cite{kipf2016variational, hu2020strategies, hu2020gpt, lu2021learning,liu2025graphpositionalautoencodersselfsupervised,jiang2023incomplete}. These approaches involve pre-training a graph encoder using self-supervised objectives, and then adapt the pre-trained knowledge to downstream tasks. However, a significant gap exists between the objectives of pre-training and those of downstream tasks, resulting in suboptimal performance. 

\stitle{Graph prompt learning.}
First proposed in NLP, prompt learning has emerged as a powerful framework for bridging the gap between pre-training and downstream tasks \cite{brown2020language,liu2021gpt,lester2021power}. Recently, this paradigm has been extended to the graph domain as a compelling alternative to fine-tuning approaches \cite{liu2023graphprompt,sun2022gppt,yu2023hgprompt,yu2024generalized,yu2023multigprompt}. These methods typically utilize a universal template to align pre-training and downstream tasks, followed by task-specific prompts that facilitate seamless adaptation to downstream tasks while keeping the pre-trained model frozen. However, current graph prompt learning methods directly produce a final answer in a single step, resulting in insufficient refinement to the answer \cite{yu2024text,yu2025samgpt,fang2025few}.

\section{Preliminaries}\label{sec.preliminaries}

In this section, we present the background and preliminaries relevant to our work.

\stitle{Graph.}
A graph is defined as \( G = (V, E) \), where \( V \) is the set of nodes and \( E \) is the set of edges. The nodes are associated with a feature matrix $\mathbf{X} \in \mathbb{R}^{|V| \times d}$, where \( \vec{x}_v \in \mathbb{R}^d \) is a row of $\mathbf{X}$ representing the feature vector for node \( v \in V \). For a collection of multiple graphs, we denote it as \( \mathcal{G} = \{ G_1, G_2, \dots, G_N \} \).

\begin{figure*}[t]
\centering
\includegraphics[width=1\linewidth]{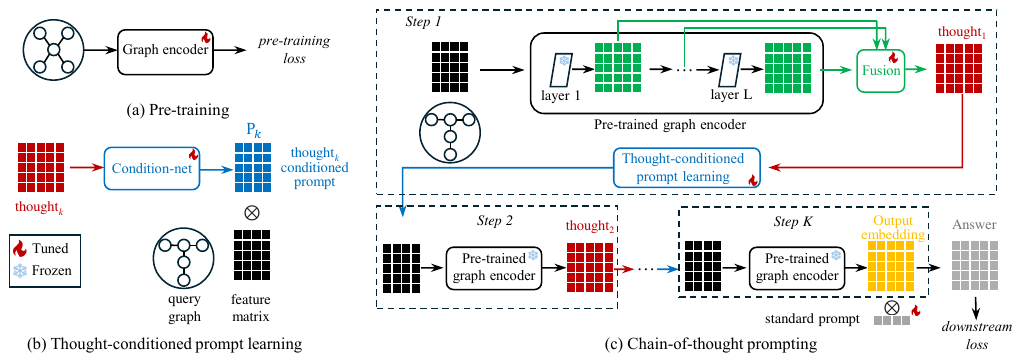}
\caption{Overall framework of \model.}
\label{fig.framework}
\end{figure*}

\stitle{Graph encoder.}
Towards graph representation learning, one of the most widely used families of graph encoders is graph neural networks (GNNs), which generally rely on message passing to capture structural knowledge \cite{wu2020comprehensive,zhou2020graph}. Each node updates its representation by aggregating information from its neighbors, and stacking multiple GNN layers enables iterative message propagation across the graph.
Formally, let $\vec{H}^l$ denote the embedding matrix at the $l$-th layer, where each row $\vec{h}_i^l$ represents the embedding of node $v_i$. This matrix is iteratively computed using the embeddings from the preceding layer:
\begin{equation}\label{eq.gnn}
\vec{H}^l = \textsc{MP}(\vec{H}^{l-1},G;\theta^l),
\end{equation}
where $\textsc{MP}(\cdot)$ is the message passing function, and $\theta^l$ represents the learnable parameters of the graph encoder at layer $l$. The initial embedding matrix, $\vec{H}^0$, is the input feature matrix, \ie, $\vec{H}^0=\vec{X}$. The output after a total of $L$ layers is then $\vec{H}^L$; for brevity we may simply write $\vec{H}$. We abstract the multi-layer encoding process as 
\begin{align}
    \{\vec{H}^1, \vec{H}^2, \cdots, \vec{H}^L\} = \textsc{GraphEncoder}(\vec{X},G;\Theta),
\end{align}
where $\{\vec{H}^1, \vec{H}^2, \cdots, \vec{H}^L\}$ denotes the embedding matrix of each layer of the graph encoder, respectively. $\Theta=(\theta^1,\ldots,\theta^L)$ is the collection of weights across the layers.

\stitle{Pre-training.}
As prior studies \cite{yu2024generalized,yu2024non} have suggested, mainstream contrastive pre-training tasks on graphs \cite{liu2023graphprompt,velickovic2019deep,you2020graph} can be unified under the task template of similarity calculation.
Formally, the unified pre-training objective is defined as follows:
\begin{equation}\label{eq:generalized_loss}
     \bL(\Theta)= -\sum_{o\in \bT_\text{pre}}\ln\frac{\sum_{a\in \text{Pos}_o}\exp(\text{sim}(\vec{h}_{a}, \vec{h}_{o})/\tau)}{\sum_{b\in \text{Neg}_o}\exp(\text{sim}(\vec{h}_{b}, \vec{h}_{o})/\tau)},
\end{equation}
where \( \text{Pos}_o \) and \( \text{Neg}_o \) denote the sets of positive and negative samples for a target instance \( o \), respectively. \( \vec{h}_o \) represents the embedding of the target instance, while \( \vec{h}_a \) and \( \vec{h}_b \) correspond to the embeddings of positive and negative samples. The hyperparameter \( \tau \) controls the temperature scaling of the similarity function $\text{sim}(\cdot,\cdot)$. In our framework, we follow previous work \cite{liu2023graphprompt,yu2024generalized} by employing similarity calculation as the task template and using link prediction as the pre-training task.

\stitle{Problem definition.}
In this work, we explore a CoT prompt learning framework for text-free graphs. We focus on two widely used tasks in graph learning: node classification and graph classification, in few-shot scenarios.
For node classification, given a graph \( G = (V, E) \) with a set of node classes \( Y \), each node \( v_i \in V \) is associated with a label \( y_i \in Y \). In contrast, graph classification considers a collection of graphs \( \mathcal{G} \), where each graph \( G_i \in \mathcal{G} \) is assigned a class label \( Y_i \in Y \).
In the few-shot setting, only \( m \) labeled examples per class are available (\eg, \( m \leq 10 \)), which is referred to as \( m \)-shot classification \cite{liu2023graphprompt,yu2024generalized}.

\section{Chain-of-Thought Graph Prompt Learning}

In this section, we present our approach, \model, starting with an overview of its framework. Then we detail its core components and conclude with a complexity analysis of the algorithm.

\subsection{Overall Framework}
We illustrate the overall framework of \model\ in Fig.~\ref{fig.framework}, which consists of two phases: pre-training and CoT prompting. 

First, we pre-train a graph encoder as shown in Fig.~\ref{fig.framework}(a). Details of the pre-training phase are provided in Sect.~\ref{sec.preliminaries}.

Second, given a pre-trained graph encoder, to guide the model with additional inference steps before finalizing its predictions, we propose CoT prompting, as illustrated in Fig.~\ref{fig.framework}(c). Specifically, we design an inference step with three substages: prompt-based inference, thought construction, and thought-conditioned prompt learning. Specifically, in each inference step, we first feed the prompt and query graph into the pre-trained graph encoder. Then, we construct a ``thought'' by fusing embeddings of all hidden layers of the pre-trained graph encoder. Lastly, as detailed in Fig.~\ref{fig.framework}(b), conditioned on the constructed thought, we employ a condition-net to generate a series of node-specific prompts. These conditional prompts capture individualized patterns for nodes in a fine-grained and parameter-efficient manner. We repeat the inference step one or more times before making the final prediction for the downstream task. It is worth noting that, to better align the downstream task with the pre-training task, we still incorporate a standard prompt similar to those used in prior graph prompting methods \cite{liu2023graphprompt,fang2024universal}.

\subsection{Chain-of-Thought Prompting}

Next, we introduce the details of each inference step and the standard prompt learning employed in our framework. 

\subsubsection{Inference step} Each inference step consists of three substages, as described in the following.

\stitle{Prompt-based inference.}
In the \(k\)-th inference step, we first feed the query graph $G=(V,E)$, along with its prompt-modified feature matrix, into the pre-trained encoder:
\begin{equation}
    \{\vec{H}_k^1, \vec{H}_k^2, \cdots, \vec{H}_k^L\} = \textsc{GraphEncoder}(\vec{X}_{k},G;\Theta_0),
\end{equation}
where $\Theta_0$ denotes the frozen pre-trained weights in the graph encoder, and $\vec{X}_{k}$ is the node feature matrix modified by the prompts generated in the previous (\ie, $k-1$-th) step. The generation of the prompts and the modification of the feature matrix are elaborated later in \textit{thought-conditioned prompt learning}.  Note that for the first step ($k=1$), the original feature matrix is used without modification, \ie, $\vec{X}_1=\vec{X}$.

\stitle{Thought construction.}
To leverage the hierarchical knowledge across multiple layers of the graph encoder, we construct a ``thought''  by fusing the hidden embeddings from each layer of the pre-trained graph encoder as follows:
\begin{equation}
    \vec{T}_{k} = \textsc{Fuse}(\vec{H}_k^1, \vec{H}_k^2, \cdots, \vec{H}_k^L),
\end{equation}
where each \(\vec{H}_k^l\) denotes the hidden embedding from the \(l\)-th layer during the \(k\)-th inference step. The $\textsc{Fuse}(\cdot)$ function can be implemented in various ways. For example, if all hidden layers of the graph encoder share the same dimensionality, we can adopt a simple weighted summation\footnote{This is the case in our experiments. If the dimensionalities differ, linear layers can be potentially used to project them to a common dimension.}, as follows.
\begin{equation}
    \vec{T}_{k} = w^1\cdot\vec{H}_k^1+w^2\cdot\vec{H}_k^2+\cdots+w^L\cdot\vec{H}_k^L,\label{eq:thought_construction}
\end{equation}
where $W=\{w^l \in \mathbb{R}:1\le l\le L\}$ are learnable parameters. 
The resulting thought, $\vec{T}_{k}$, captures the current working state of the graph encoder, storing multi-layer structural knowledge. In particular, the $i$-th row of $\vec{T}_{k}$ reflects the thought pertaining specifically  to node $v_i$. 

\stitle{Thought-conditioned prompt learning.}
The thought constructed in Eq.~\eqref{eq:thought_construction} is then used for prompt generation, in order to further guide the next inference step. Moreover, since different nodes may exhibit distinct characteristics with respect to the downstream task,  it is beneficial to adapt the pre-trained model to node-specific patterns. 
Thus, rather than employing a single prompt for all nodes, we propose to leverage a conditional network (condition-net) \cite{zhou2022conditional,yu2025node,yu2024non} to generate node-specific prompts and facilitate node-specific adaptation. Specifically, conditioned on the current thought \(\vec{T}_{k}\), the condition-net generates a series of prompts, collectively represented as $\vec{P}_{k}\in \mathbb{R}^{|V|\times d}$, as follows.
\begin{align}\label{eq.prompt-generation}
    \vec{P}_{k} = \textsc{CondNet}(\vec{T}_{k}; \phi),
\end{align}
where \(\textsc{CondNet}\) is the condition-net parameterized by \(\phi\). The condition-net can be viewed as a hypernetwork \cite{ha2022hypernetworks}---a lightweight auxiliary network such as a multi-layer perceptron (MLP)---that generates a distinct prompt for each node from the thought, yet avoids parameterizing a separate prompt vector for each node. 
In particular, the $i$-th row of \(\vec{P}_{k}\) represents a unique node-specific prompt vector \( \vec{p}_{k,i} \) for node $v_i \in V$. 
Subsequently, the prompt vectors are used to modify the node features of the query graph for \emph{prompt-based inference} in the next (\ie, $k+1$-th) step, as follows.
\begin{align}\label{eq.thought-prompting}
    \vec{X}_{k+1} = \vec{P}_k \odot \vec{X},
\end{align}
where \(\odot\) denotes element-wise multiplication, and \(\vec{X}_{k+1}\) represents the input to the pre-trained graph encoder in the \(k+1\)-th step. 

\subsubsection{Standard prompt learning} 
To align the objectives of pre-training and downstream tasks, we also leverage a standard prompt following prior work, which typically modifies node features \cite{sun2022gppt,fang2024universal,yu2024generalized} or embeddings \cite{liu2023graphprompt,yu2024generalized}. 
In particular, we adopt GPF+ \cite{fang2024universal} to generate the standard prompts. However, we emphasize that our framework is compatible with any standard graph prompting technique. As demonstrated in Sect.~\ref{sec.backbone-flexibility}, our CoT-style prompting can be combined with several prevailing graph prompting approaches to further enhance their performance. 

 Specifically, we train \( N \) bias prompts $\{\vec{p}_\text{bias}^1, \ldots, \vec{p}_\text{bias}^N\}$ 
and leverage attention-based aggregation to generate node-specific prompts. While GPF+ \cite{fang2024universal} applies these prompts to graph features $\vec{X}$, we use them to modify the output embeddings $\vec{H}_K$ after $K$ inference steps. 
Concretely, the standard prompt for node \( v_i \) is computed as
\begin{equation}
    \vec{p}_{\text{std},i} = \sum_{j=1}^{N} \alpha_{i,j} \vec{p}_\text{bias}^j,\quad \text{where} \
    \alpha_{i,j} = \frac{\exp\big(\vec{a}^{j}\vec{h}_{K,i}\big)}{\sum_{n=1}^{N} \exp\big(\vec{a}^{n}\vec{h}_{K,i}\big)}.
\end{equation}
Here, \( \vec{h}_{K,i} \) denotes the \( i \)-th row of the output embedding matrix \( \vec{H}_{K} \), \ie, the embedding vector of node \( v_i \) after $K$ inference steps. \(\{\vec{a}^1, \vec{a}^2, \ldots, \vec{a}^N\} \) are \( N \) learnable linear projection vectors. The standard prompts for all nodes are stacked to form the matrix \( \vec{P}_\text{std} \), which is then used to modify the output embeddings $\vec{H}_{K}$:
\begin{equation}\label{eq.initial-prompt}
    \vec{\Tilde{H}} = \vec{P}_\text{std} \odot \vec{H}_{K}.
\end{equation}
We call $\vec{\Tilde{H}}$ the \emph{answer matrix}, since it is used by downstream tasks to produce final predictions (or answers) in Sect.~\ref{sec.prompt-tuning}.

Note that for clarity and to maintain a unified representation of various standard prompt mechanisms, we use  \(\vec{P}_\text{std}\) to denote the trainable parameters associated with these mechanisms throughout the remainder of this paper. 

\subsubsection{Summary} 
In \model, we adopt the following mechanisms, which collectively ensure the effectiveness of \model. (1) We follow standard graph prompt learning methods \cite{liu2023graphprompt,fang2024universal} to align pre-training and downstream tasks, ensuring that \model\ can efficiently adapt to different downstream tasks even in few-shot settings. (2) The CoT-style prompting performs multiple inference steps, allowing iterative refinement of its prediction for a given downstream task. (3) The thoughts in \model\ fuse hierarchical topological knowledge from graphs, enabling the capture of fine-grained structural information. (4) Conditioned on the thought, \model\ generates a series of node-specific prompts that reflect individualized node characteristics in a parameter-efficient manner.

\subsection{Prompt Tuning}\label{sec.prompt-tuning}

Consider a downstream task with a labeled training set 
\[
\mathcal{D} = \{(x_1, y_1), (x_2, y_2), \dots\},
\]
where each \( x_i \) represents either a node or a graph and \( y_i \in Y \) denotes its corresponding class label. Subsequently, the loss function of the task is defined as $\bL_{\text{down}}(\mathcal{D};W,\phi,\vec{P}_\text{std})=$
\begin{align}\label{eq.prompt-loss}
     -\sum_{(x_i, y_i) \in \mathcal{D}} \ln \frac{\exp\left(\text{sim}\left(\vec{\tilde{h}}_{x_i}, \vec{\tilde{h}}_{y_i}\right)/\tau\right)}{\sum_{c \in Y} \exp\left( \text{sim}\left(\vec{\tilde{h}}_{x_i}, \vec{\tilde{h}}_{c}\right)/\tau\right)},
\end{align}
where \( \vec{\tilde{h}}_{x_i} \) represents the final embedding of a node \( v \) or a graph \( G \). Specifically, for node classification, \( \vec{\tilde{h}}_{v} \) corresponds to a row in the answer matrix \( \vec{\tilde{H}} \); for graph classification, we apply a readout operation and compute the graph embedding as $\vec{\tilde{h}}_{G} = \sum_{v \in V} \vec{\tilde{h}}_{v}$. Lastly,
\( \vec{\tilde{h}}_{c} \) denotes the prototype embedding for class \( c \), which is obtained by averaging the embeddings of all labeled nodes or graphs belonging to that class.

During prompt tuning, only the weights of thought construction ($W$) and the condition-net (\( \phi \)), as well as the standard prompts (\(\vec{P}_\text{std}\)), are updated, while the pre-trained weights of the graph encoder remain frozen. This parameter-efficient design makes our approach well-suited for few-shot learning, where the training set \( \mathcal{D} \) contains only a few labeled examples.

\subsection{Algorithm and Complexity Analysis}

\begin{algorithm}[tbp]
\small
\caption{\textsc{Chain-of-Thought Graph Prompt Learning}}
\label{alg.prompt}
\begin{algorithmic}[1]
    \Require Pre-trained graph encoder with parameters $\Theta_0$, labeled data $\mathcal{D}$ for a given downstream task
    \Ensure Optimized parameters $W, \phi, \vec{P}_{\text{std}}$.
    \State $W,\phi, \vec{P}_{\text{std}} \leftarrow$ initialization
    \While{not converged} 
        \State $\vec{X}_1\leftarrow \vec{X}$
        \State \slash* First $K-1$ inference steps*\slash
        \While{inference step $1\leq k< K$}
            \State \slash* Prompt-based inference *\slash
            \State $\{\vec{H}_k^1, \vec{H}_k^2, \cdots, \vec{H}_k^L \}\leftarrow \textsc{GraphEncoder}(G,\vec{X}_k;\Theta_0)$
            \State \slash* Thought construction *\slash
            \State $\vec{T}_{k} \leftarrow \mathtt{Fuse}(\vec{H}_k^1, \vec{H}_k^2, \cdots, \vec{H}_k^L)$
            \State \slash*  Thought-conditioned prompt generation by Eq.~\eqref{eq.prompt-generation} *\slash
            \State $\vec{P}_{k} \leftarrow \mathtt{CondNet}(\vec{T}_{k}; \phi)$
            \State \slash* Feature modification by Eq.~\eqref{eq.thought-prompting} *\slash
            \State $\vec{X}_{k+1} \leftarrow \vec{P}_{k} \odot \vec{X}$
        \EndWhile
            \State \slash* Last (\ie, $K$-th) inference step *\slash
            \State $\{\vec{H}_K^1, \vec{H}_K^2, \cdots, \vec{H}_K^L\} \leftarrow \textsc{GraphEncoder}(G,\vec{X}_K;\Theta_0)$
            \State $\vec{H}_{K}\leftarrow \vec{H}_K^L$
            \State \slash* Standard prompt modification by Eq.~\eqref{eq.initial-prompt} *\slash
            \State $\vec{\tilde{H}} \leftarrow \vec{P}_\text{std} \odot \vec{H}_{K}$
            \State \slash* Update prototypical instance *\slash
            \For{each class $c$} 
                \State ${\vec{\tilde{h}}}_{c} \leftarrow \textsc{Mean}(\{\vec{\tilde{h}}_{x}$: $x$ is any instance in  class $c$ from $\mathcal{D}$\})
            \EndFor
            \State \slash* Optimizing the parameters *\slash
            \State Calculate $\bL_\text{down}(\mathcal{D};W,\phi,\vec{P}_\text{std})$ by Eq.~\eqref{eq.prompt-loss}
            \State Update $W,\phi,\vec{P}_\text{std}$ by backpropagating  $\bL_\text{down}(\mathcal{D};W,\phi,\vec{P}_\text{std})$
        \EndWhile    
    \State \Return $\{W,\phi,\vec{P}_\text{std}\}$
\end{algorithmic}
\end{algorithm}

\stitle{Algorithm.}
We outline the main steps for \model\ in Algorithm~\ref{alg.prompt}. In lines 3--13, we iterate through the first \( K-1 \) inference steps for a downstream task while keeping the pre-trained weights \( \Theta_0 \) frozen. Specifically, in lines 8--9, we generate the thought for the \( k \)-th inference step by fusing the  embeddings from each layer of the pre-trained graph encoder. In lines 10--11, we leverage this thought to generate node-specific prompts that guide the next inference step. In lines 14--18, we perform the last inference step,  employing a standard prompt to modify the output embedding and obtaining the answer matrix. 
Finally, in lines 19--24, we calculate and backpropagate the downstream loss to update the parameters $W,\phi,\vec{P}_\text{std}$.

\stitle{Complexity analysis.}
For a downstream graph \( G \), we perform \( K \) inference steps. In each step, we first conduct prompt-based inference, the complexity of which is determined by the architecture of the graph encoder. In a standard GNN, each node aggregates information from up to \( D \) neighboring nodes per layer. Consequently, computing node embeddings over \( L \) layers results in a complexity of \( O(D^L  |V|) \), where \( |V| \) denotes the number of nodes. This is followed by thought generation, which fuses embeddings from all \( L \) layers, introducing an additional complexity of \( O(L  |V|) \). The generated thought is subsequently used for thought-conditioned prompt learning, with a complexity of \( O(|V|) \). Therefore, the computational complexity for the \( K \) inference steps is \( O(K  (D^L + L)  |V|) \).
Additionally, we employ a standard prompt to modify node features or embeddings. The complexity of this process depends on the specific prompting mechanism. Here, we assign it a complexity of \(O(|V|)\), which is common among graph prompting methods \cite{liu2023graphprompt,fang2024universal}.
Thus, the overall complexity of \model\ is $O(K (D^L+L)  |V| )$, or more simply $O(K D^L  |V| )$, since it is typical that $L \ll D^L$.

\section{Experiments}
In this section, we conduct experiments to evaluate \model, and analyze the empirical results.

\begin{table}[tbp]
\center
\small
\addtolength{\tabcolsep}{-1mm}
\caption{Summary of datasets. 
\label{table.datasets}}
\resizebox{1\columnwidth}{!}{%
\begin{tabular}{@{}c|rrrrrrc@{}}
\toprule
    Datasets & \makecell[c]{Graphs} & \makecell[c]{Graph \\ classes} & \makecell[c]{Avg.\\ nodes} & \makecell[c]{Avg. \\ edges} & \makecell[c]{Node \\ features} & \makecell[c]{Node \\ classes} & \makecell[c]{Task$^*$ \\ (N/G)}\\
\midrule
     Cora & 1 & - & 2,708 & 5,429 & 1,433 & 7 & N\\ 
     Citeseer & 1 & - & 3,327 & 4,732 & 3,703 & 6 & N\\ 
     Pubmed & 1 & - & 19,717 & 88,648 & 500 & 3 & N\\
     Photo & 1 & - & 7,650 & 238,162 & 745 & 8 & N\\
     MUTAG & 188 & 2 & 17.9 & 18.9 & 7 & - & G\\
     COX2 & 467 & 2 & 41.2 & 43.5 & 3 & - & G\\
     BZR & 405 & 2 & 35.8 & 38.4 & 3 & - & G\\
     PROTEINS & 1,113 & 2 & 39.1 & 72.8 & 4 & 3 & G\\
 \bottomrule
\end{tabular}}\\[1mm]
\parbox{1\columnwidth}{\raggedright \footnotesize \( ^* \) This column indicates the type of downstream task conducted for each dataset: ``N'' denotes node classification, while ``G'' denotes graph classification.}
\end{table}

\begin{table*}[tbp] 
    \centering
    \small
    \caption{Accuracy (\%) evaluation of node and graph classification.
    }
    \label{table.node-graph-classification}%
    \begin{tabular}{l|cccc|cccc}
    \toprule
  \multirow{2}*{Methods} & \multicolumn{4}{c|}{Node classification} & \multicolumn{4}{c}{Graph classification} \\ 
  & Cora  & Citeseer & Pubmed & Photo & MUTAG & COX2 & BZR & PROTEINS \\\midrule\midrule
    \method{GCN} & 32.50 \text{± 14.21}  & 26.36 \text{± \ \ 9.03}  & 52.18 \text{± \ \ 8.70}  & 60.18 \text{± 12.04}
    &  43.44 \text{± 15.14}  & 50.95 \text{± 23.48}  & 47.25 \text{± 16.59} & 40.28 \text{± \ \ 0.03} \\
    \method{GAT} & 31.00 \text{± 16.22}  & 27.71 \text{± \ \ 8.74}  & 50.02 \text{± \ \ 8.88}  & 51.79 \text{± 12.85}
    &  37.33 \text{± 10.81}  & 50.58 \text{± 26.16}  & 46.55 \text{± 16.57}  & 40.39 \text{± \ \ 0.04}\\
    \midrule
    \method{DGI/InfoGraph} & 54.11 \text{± \ \ 9.60}  & 45.00 \text{± \ \ 9.19}  & 47.46 \text{± 12.19}  & 58.89 \text{± 10.97} 
    &  53.17 \text{± 17.29} & 53.82 \text{± 14.19}  & 49.33 \text{± 15.11}  & 52.51 \text{± 10.29}\\
    \method{GraphCL} & 51.96 \text{± \ \ 9.43}  & 43.12 \text{± \ \ 9.61}  & 46.80 \text{± \ \ 9.04}  & 57.78 \text{± 11.31}
    &  54.92 \text{± 17.09}  & 53.81 \text{± 14.21}  & 49.73 \text{± 14.66}  & 53.81 \text{± \ \ 8.97}\\
    \midrule
    \method{ProG} & 50.59 \text{± 14.64} & 43.17 \text{± \ \ 8.49}  & 63.07 \text{± 11.96}  & 66.50 \text{± \ \ 9.46}
    & 51.99 \text{± \ \ 4.50} & 53.45 \text{± 15.01}  & 53.52 \text{± 11.97}  & 52.73 \text{± \ \ 6.57}\\
    \method{GPF} & \underline{57.60} \text{± 13.88}  & 43.11 \text{± \ \ 8.80}  & 55.63 \text{± 10.96}  & 65.29 \text{± 10.07}
    &  56.55 \text{± 13.95}  & 54.16 \text{± 14.07}  & 48.65 \text{± 13.96} & 53.05 \text{± \ \ 7.62}\\
    \method{GPF+} & 57.42 \text{± 13.87}  & 43.28 \text{± \ \ 8.82}  & 57.16 \text{± 10.99}  & 65.07 \text{± 10.01}
    &  \underline{56.81} \text{± 12.93}  & \underline{55.24} \text{± 13.29}  & 50.83 \text{± 19.74} & \underline{54.58} \text{± \ \ 8.70} \\
    \method{GraphPrompt} & 54.25 \text{± \ \ 9.38}  & \underline{45.34} \text{± 10.53} & \underline{63.11} \text{± 10.01}  & \underline{66.62} \text{± \ \ 9.90}
    &  55.44 \text{± 12.56}  & 54.34 \text{± 14.77}  & \underline{54.59} \text{± 10.52}  & 53.80 \text{± \ \ 7.93}\\
    \midrule
    \model & \textbf{59.67} \text{± 15.51}  & \textbf{46.21} \text{± \ 8.78}  & \textbf{64.43} \text{± \ 9.96}  & \textbf{67.16} \text{± 10.46} 
    & \textbf{58.75} \text{± 15.42}  & \textbf{56.26} \text{± 15.52}  & \textbf{58.03} \text{± 23.44}  & \textbf{56.24} \text{± \ 8.60}\\
    \bottomrule
        \end{tabular}\\[1mm]
       \parbox{1\linewidth}{\footnotesize \ \ \ \ \ \ Best results are \textbf{bolded} and runner-up results are \underline{underlined}.}
\end{table*}

\subsection{Experimental Setup}\label{sec.exp-setup}
\stitle{Datasets.}
We conduct experiments on eight widely used benchmark datasets, spanning citation networks, e-commerce, protein structures, and molecular graphs.
\textit{Cora} \cite{mccallum2000automating}, \textit{Citeseer} \cite{sen2008collective}, and \textit{Pubmed} \cite{sen2008collective} are citation networks, each consisting of a single graph. In these datasets, nodes represent academic papers and edges denote citation relationships between them.
\textit{Photo} \cite{shchur2018pitfalls} is an e-commerce co-purchase network derived from Amazon's photography-related product categories, where nodes correspond to products and edges indicate frequently co-purchased items.
\textit{PROTEINS} \cite{borgwardt2005protein} is a dataset of protein structures. In each graph, nodes correspond to secondary structures, and edges capture spatial or sequential relationships within the amino acid sequence.
\textit{MUTAG} \cite{nr}, \textit{BZR} \cite{nr}, and \textit{COX2} \cite{nr} are molecular graph datasets, representing nitroaromatic compounds, ligands associated with benzodiazepine receptors, and  molecular structures related to cyclooxygenase-2 inhibitors, respectively.
These datasets are summarized in Table~\ref{table.datasets}, with further descriptions in Appendix~\ref{app.dataset}.

\stitle{Baselines.}
We compare \model\ with state-of-the-art methods across three categories.
(1) \emph{Supervised GNNs}: GCN \cite{kipf2016semi} and GAT \cite{velivckovic2017graph} are trained directly on downstream labels in a supervised manner, without any pre-training.
(2) \emph{Graph pre-training models}: DGI/InfoGraph\footnote{DGI is originally designed for node-level tasks, while InfoGraph extends it to graph-level classification. In our experiments, we apply DGI to node classification and InfoGraph to graph classification.} \cite{velivckovic2017graph,sun2019infograph} and GraphCL \cite{you2020graph} adopt a ``pre-train, fine-tune'' strategy. They first perform self-supervised pre-training using unlabeled graphs and are later fine-tuned for downstream tasks, where a classifier is trained with few-shot labels while keeping the pre-trained encoder frozen. 
(3) \emph{Graph prompt learning models}: ProG \cite{sun2023all}, GPF \cite{fang2024universal}, GPF+ \cite{fang2024universal}, and GraphPrompt \cite{liu2023graphprompt} employ self-supervised pre-training followed by prompt tuning. Unlike the fine-tuning methods, these methods leverage a unified task template, and train task-specific prompts for downstream adaptation. 
Further descriptions of these baselines are presented in Appendix~\ref{app.baselines}.


\stitle{Downstream tasks and evaluation.}
We perform experiments on two downstream tasks: node classification and graph classification. Both types of task follow an \( m \)-shot classification setup, where for each class, we randomly select \( m \) instances (nodes or graphs) as labeled examples. The remaining instances are treated as the test set.
In our main results, we set \( m = 1 \) for both node and graph classification tasks. Additionally, to examine the robustness of our method, we vary \( m \) within the range \( [1, 10] \), allowing us to analyze the performance under different few-shot scenarios.
We construct 100 independent \( m \)-shot tasks for each type of classification through repeated sampling. Each task is evaluated using five different random seeds, resulting in a total of 500 experimental runs per task type. We adopt accuracy as the performance metric and report both the mean and standard deviation across these runs, in line with previous studies \cite{wang2020graph,liu2021relative,liu2023graphprompt}.

\subsection{Implementation Details} \label{app.parameters}

We outline key settings for the baselines and \model. 

\stitle{Baseline settings.}
We utilize the official codes for all open-source baselines. Each model was tuned based on the settings recommended in their respective work to achieve optimal performance.

For the baseline GCN \cite{kipf2016semi}, we employ a 2-layer architecture, and set the hidden dimensions to 64. 
For GAT \cite{velivckovic2017graph}, we employ a 2-layer architecture and set the hidden dimension to 64. Additionally, we apply 8 attention heads in the first GAT layer.
For DGI \cite{velivckovic2017graph}, we utilize a 1-layer GCN as the backbone and set the hidden dimensions to 256. Additionally, we employ PReLU as the activation function.
For InfoGraph \cite{sun2019infograph}, a 1-layer GCN is used as the backbone, with its hidden dimensions set to 256.
For GraphCL \cite{you2020graph}, a 1-layer GCN is also employed as its backbone, with the hidden dimensions set to 256. Specifically, we select edge dropping as the augmentations, with a default augmentation ratio of 0.2.
For GraphPrompt \cite{liu2023graphprompt}, a 3-layer GCN is used as the backbone for all datasets, with the hidden dimensions set to 256.
For GPF and GPF+ \cite{fang2024universal}, we employ a 3-layer GCN as the backbone for all datasets. The hidden dimensions are set to 256.
For ProG \cite{sun2023all}, we employ a 2-layer GCN as the backbone for all datasets. The hidden dimensions are set to 100.

\stitle{\model\ settings.}
For our proposed \model, we utilize a 3-layer GCN as the backbone for all datasets, with the hidden dimensions set to 256. We set the number of inference step as 2 for node classification task, and 3 for graph classification task. The condition-net is implemented as a two-layer MLP with a bottleneck architecture \cite{yu2024non,yu2025node}. Its input dimension is 256, while the hidden dimension is set to 32 for node classification and 8 for graph classification.

\stitle{Environment.}
\label{app.general-setting}
All experiments were conducted on Ubuntu 22.04.2, using a machine equipped with AMD EPYC 7742 64-core processors and NVIDIA GeForce RTX 3090 (24 GB) GPUs.

\subsection{Performance Evaluation}\label{sec.exp.per}
We first evaluate one-shot classification tasks. Then, we examine the model performance as the number of shots increases. Note that for the other experiments in Sect.~\ref{sec.ablation} and thereafter, we adopt the one-shot setting.

\stitle{One-shot performance.}\label{exp.main}
We present the results for one-shot node and graph classification in Table~\ref{table.node-graph-classification}. We make several major observations, as follows.

(1) 
\model\ consistently outperforms the baseline methods across both node and graph classification, demonstrating its overall advantage and robustness.

(2) Supervised methods tend to underperform compared to others, as they do not leverage any pre-trained model. On the other hand, graph pre-training models achieve improved performance through self-supervised pre-training on unlabeled graphs. 

(3) Standard graph prompt learning approaches---ProG, GPF, GPF+, and GraphPrompt---often outperform fine-tuning of the pre-trained models, due to their alignment between pre-training and downstream objectives, as well as their  parameter efficiency. However, they still fall short compared to \model, due to their single-step inference process that lacks iterative refinement of the final answer. This highlights the effectiveness of our CoT-style prompting, which enables step-by-step inference.

\begin{figure}[t]
\centering
\includegraphics[width=1\linewidth]{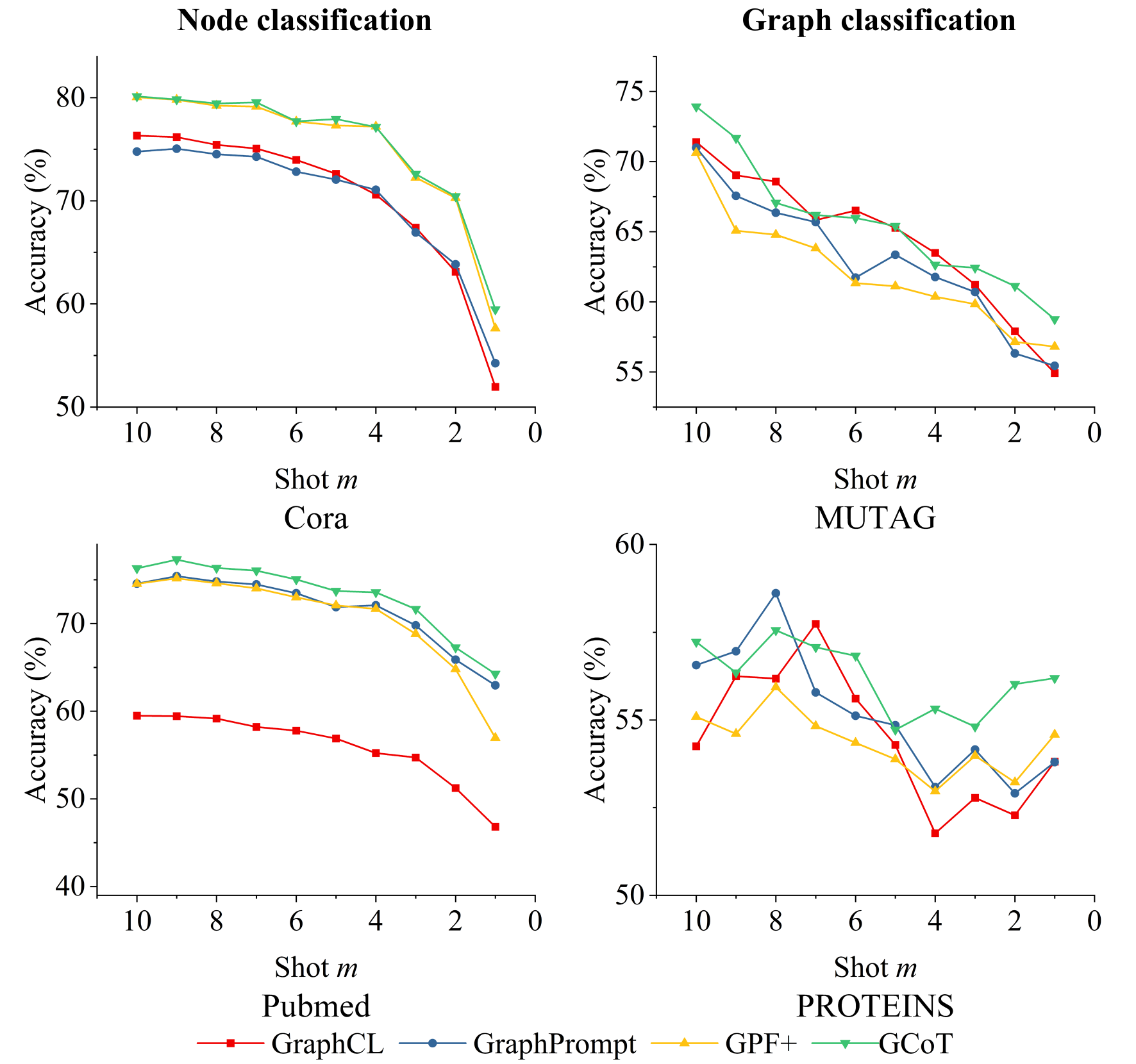}
\caption{Impact of labeled data size (number of shots) on node and graph classification.}
\label{fig.fewshot}
\end{figure}

\stitle{Few-shot performance.}
To investigate the effect of labeled data size on performance, we vary the number of shots in both node and graph classification tasks. The results are presented in Fig.~\ref{fig.fewshot}, comparing \model\ against several competitive baselines. We make the following observations.

(1) \model\ generally outperforms the baselines on both node and graph classification tasks, particularly in low-shot settings (\eg, \(m \leq 5\)), where labeled data are limited.


(2) As the number of shots increases (\eg, \( m > 5 \)), all methods generally exhibit improved performance, as expected. Nevertheless, \model\ remains highly competitive, often achieving the best or near-best results.

(3) On the \emph{PROTEINS} dataset, the performance of all methods tends to fluctuate more. A possible reason is that this dataset exhibits greater variability in graph sizes compared to other datasets: The standard deviation of graph sizes in \emph{PROTEINS} is 45.78, whereas other datasets fall within the range of 4.04 to 15.29. This may contribute to the unstable performance.
Despite this, \model\ demonstrates greater robustness than the competing methods.

\subsection{Ablation Study and Visualization}\label{sec.ablation}

To thoroughly evaluate the contribution of CoT-style prompting to graph models, we perform an ablation study and visualization. 

\stitle{Ablation study.}
To investigate the effect of step-by-step inference and the impact of thoughts constructed from different layers, we compare \model\ with four  variants.
The first variant is \model$\backslash$CoT, which produces the final answer in a single inference step without CoT-style prompting.
The other three variants are \model-L1, \model-L2, and \model-L3, which utilize only the hidden embeddings from the first, second, and third layers of the pre-trained graph encoder as the thought, respectively. (Recall that we employ a 3-layer GCN as the graph encoder.)

As shown in Table~\ref{table.ablation}, \model\ consistently outperforms all variants. Specifically, the advantage over \model$\backslash$CoT underscores the importance of step-by-step inference, which enables iterative refinement of the predictions. Moreover, the advantage over single-layer-based thoughts demonstrates the effectiveness of fusing hierarchical knowledge from different layers of the pre-trained graph encoder in constructing the thoughts.

\begin{figure}[t]
\centering
\includegraphics[width=1\linewidth]{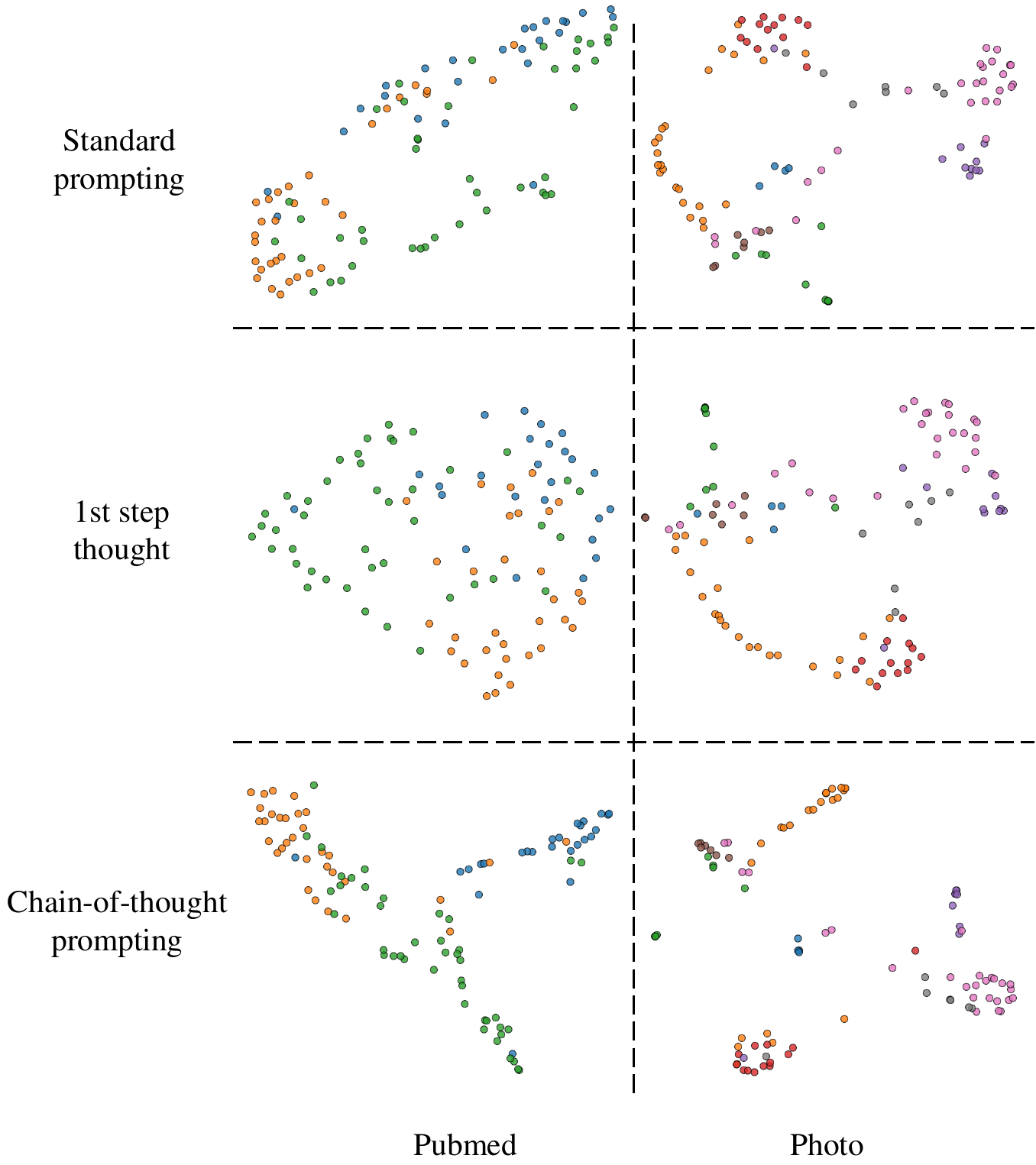}
\caption{Visualization of node and thought embeddings. Different colors represent different node classes.}
\label{fig.visualization}
\end{figure}

\stitle{Visualization.} To further demonstrate the impact of our CoT design, we visualize the output embeddings produced by \model$\backslash$CoT (the first variant in the ablation study) and the answer embeddings by \model, as well as the embeddings of thoughts constructed in the first inference step.
As shown in Fig.~\ref{fig.visualization}, each point represents the embeddings associated with a node, 
and different colors represent different classes of the nodes. 
With CoT-style prompting, node embeddings from different classes exhibit a clearer separation compared to those produced by \model$\backslash$CoT, underscoring the effectiveness of \model\ in enhancing class distinction. Moreover, the thoughts generated in the first inference step already show some clustering---though not as well as the final node embeddings produced by \model---since they only reflect an intermediate state before the final step (two steps in total are used).


\begin{table}[tbp] 
    \centering
    \small
    \caption{Ablation study on the effects of key components.
    }
    \label{table.ablation}%
    \begin{tabular}{l|cc|cc}
    \toprule
  \multirow{2}*{Methods} & \multicolumn{2}{c|}{Node classification} & \multicolumn{2}{c}{Graph classification}\\
  & {Cora} & {Pubmed} & {MUTAG} & {PROTEINS} \\\midrule\midrule
    \model$\textbackslash$CoT & 56.65\text{±13.97}& 62.80\text{±10.08}& 56.49\text{±16.61}  & 53.40\text{±6.66} \\
    \model-L1 & 57.18\text{±14.34}& 63.31\text{±10.05}& 56.54\text{±14.12} & 54.71\text{±8.57}\\
    \model-L2 & 57.00\text{±14.48}& 63.20\text{±10.08}& 57.68\text{±13.84} & 54.77\text{±8.81}\\
    \model-L3 & 57.01\text{±14.66}& 63.33\text{±10.05}& 57.85\text{±16.10}  & 56.22\text{±8.45}\\
    \model & \textbf{59.67}\text{±15.51}& \textbf{64.43}\text{± 9.96}  & \textbf{58.75}\text{±15.42}  & \textbf{56.24}\text{±8.60} \\
    \bottomrule
        \end{tabular}
\end{table}

\subsection{Heterophily Sensitivity}\label{sec.hetero}
To examine the robustness of \model\ on heterophilic graphs \cite{yu2024non}, we conduct one-shot node classification on \textit{Wisconsin} \cite{pei2020geom} and \textit{Squirrel} \cite{rozemberczki2021multi}. 
Specifically, \emph{Wisconsin} is a network of 251 nodes representing webpages, with 199 edges indicating hyperlink connections among them. Node features are constructed using a bag-of-words approach based on webpage content. 
\textit{Squirrel} \cite{rozemberczki2021multi} consists of 5,201 Wikipedia pages discussing predefined topics. Nodes represent individual webpages, while edges capture 217,073 hyperlink connections between them. 
Node features are constructed using key informative nouns extracted from the text content of the Wikipedia pages.
As shown in Table~\ref{table.hetero}, \model\ outperforms other competitive baselines on heterophilic node classification. These results indicate that our CoT-style prompting can generalize across both homophilic and heterophilic graphs.

\begin{table}[tbp] 
    \centering
    \small
     \addtolength{\tabcolsep}{1mm}
    \caption{Accuracy (\%) evaluation of node classification on heterophilic graphs.
    }
    \label{table.hetero}%
    \begin{tabular}{l|cc}
    \toprule
  {Methods}   & Wisconsin & Chameleon  \\\midrule\midrule
    DGI & 28.04 $\pm$ \phantom{0}6.47 & 19.33 $\pm$ 4.57 \\
    GraphCL & 29.85 $\pm$ \phantom{0}8.46 & \underline{27.16} $\pm$ 4.31 \\\midrule
    ProG & 28.99 $\pm$ 11.14 & 25.90 $\pm$ 3.95 \\
    GPF & \underline{32.18} $\pm$ 11.36 & 25.98 $\pm$ 3.71 \\
    GPF+ & 32.02 $\pm$ 11.17 & 26.12 $\pm$ 3.75 \\
    GraphPrompt & 31.48 $\pm$ \phantom{0}5.18 & 25.36 $\pm$ 3.99 \\\midrule
    \model & \textbf{32.80} $\pm$ 11.59 & \textbf{27.96} $\pm$ 3.89 \\
    \bottomrule
        \end{tabular}
\end{table}

\subsection{Flexibility of Graph Prompting Methods} \label{sec.backbone-flexibility}
To evaluate the flexibility and robustness of \model, we evaluate its performance using various standard graph prompting methods as the standard prompt. We integrate \textsc{ProG} \citep{sun2023all}, \textsc{GPF} \citep{fang2024universal}, \textsc{GPF+} \citep{fang2024universal}, and \textsc{GraphPrompt} \citep{liu2023graphprompt} into our framework. Specifically, \textsc{ProG}, \textsc{GPF}, and \textsc{GPF+} are employed to modify the input features at the first inference step, whereas \textsc{GraphPrompt} is utilized to modify the output embeddings at the final step.
The results for both node and graph classification on four datasets are presented in Table~\ref{table.backbone}. Across all cases, \model\ outperforms its standard prompting counterparts, highlighting the robustness and flexibility of our CoT design.

\begin{table}[tbp]
    \centering
    \small
    \caption{Accuracy (\%) evaluation of \model\ with different standard prompting methods. }
    \label{table.backbone}%
    \addtolength{\tabcolsep}{-0.15mm}
    \begin{tabular}{@{}l|cc|cc@{}}
    \toprule
    \multirow{2}*{Prompting} & \multicolumn{2}{c|}{Node classification} & \multicolumn{2}{c}{Graph classification}\\
    & {Cora} & {Pubmed} & {MUTAG} & {PROTEINS} \\\midrule\midrule
    \textsc{ProG}  & 50.59\text{±14.64}  & 63.07\text{±11.96}   & 51.99\text{±14.50 }  & 52.73\text{±6.57}   \\
    with \model & \textbf{52.62}\text{±15.27}   & \textbf{64.01}\text{±11.27}   & \textbf{58.42}\text{±15.39}  & \textbf{55.82}\text{±8.79}  \\\midrule
    \textsc{GPF}   & 57.60\text{±13.88}   & 55.63\text{±10.96}   & 56.55\text{±13.95}   & 53.05\text{±7.62}  \\
    with \model & \textbf{58.98}\text{±15.28}   & \textbf{62.63}\text{±10.01}  & \textbf{57.66}\text{±13.11}   & \textbf{56.98}\text{±8.30}  \\\midrule
    \textsc{GPF+} & 57.42\text{±13.87 }  & 57.16\text{±10.99}  & 56.81\text{±12.93}   & 54.58\text{±8.70}   \\
    with \model & \textbf{59.35}\text{±15.37}  & \textbf{62.18}\text{±10.49}   & \textbf{58.54}\text{±13.29}   & \textbf{55.45}\text{±9.20}\\\midrule
    \textsc{GraphPrompt} & 54.25\text{± 9.38}   & 63.11\text{±10.01}   & 55.44\text{±12.56}   & 53.80\text{±7.93}   \\
    with \model & \textbf{58.40}\text{±14.74}   & \textbf{63.79}\text{± 9.40}   & \textbf{58.28}\text{±15.47}   & \textbf{55.32}\text{±8.43}  \\
    \bottomrule
    \end{tabular}\\[1mm]
     \parbox{1\linewidth}{\footnotesize In each group, the first row represents a standard prompting method, and the second row represents the standard prompting coupled with our CoT design.}
\end{table}

\subsection{Hyperparameter Analysis}

Next, we study the sensitivity of two key hyperparameters, namely, the hidden dimension of the condition-net, and  the number of inference steps performed.

\begin{figure}[t]
\centering
\includegraphics[width=1\linewidth]{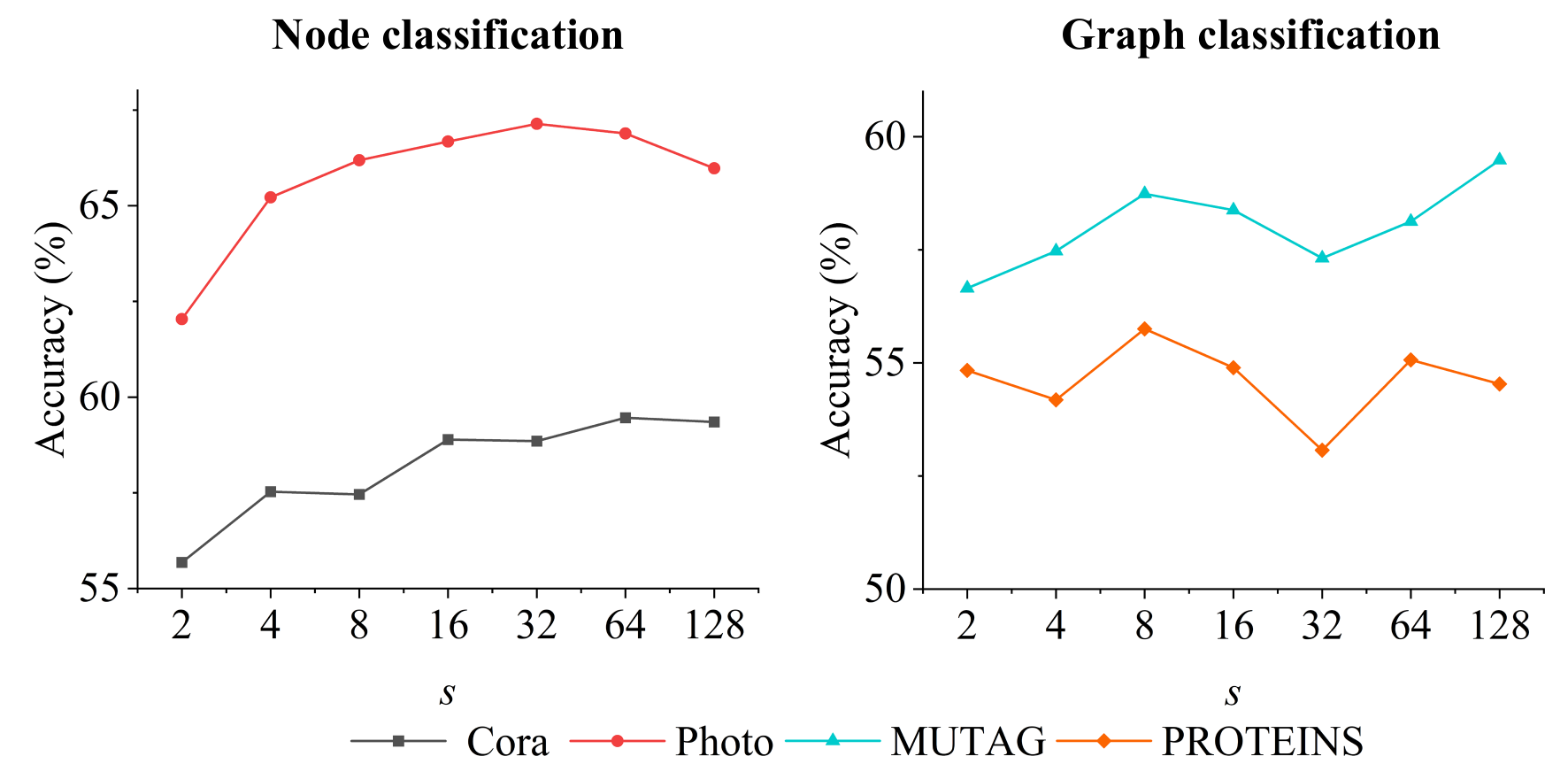}
\caption{Impact of hidden dimension $s$ in the condition-net.}
\label{fig.hiddendim}
\end{figure}

\stitle{Hidden dimension of condition-net.}
In our experiments, we implement the condition-net as a two-layer MLP with a bottleneck design. To examine the effect of the MLP architecture, we vary its hidden dimension \( s \) and present the results in Fig.~\ref{fig.hiddendim}. The results reveal that \( s = 32 \) generally yields strong results for node classification, and \( s = 8 \) for graph classification. A smaller \( s \) may restrict the model's representational capacity, limiting its effectiveness, whereas a larger \( s \) introduces additional trainable parameters and increases the risk of overfitting in few-shot scenarios.

\begin{figure}[t]
\centering
\includegraphics[width=1\linewidth]{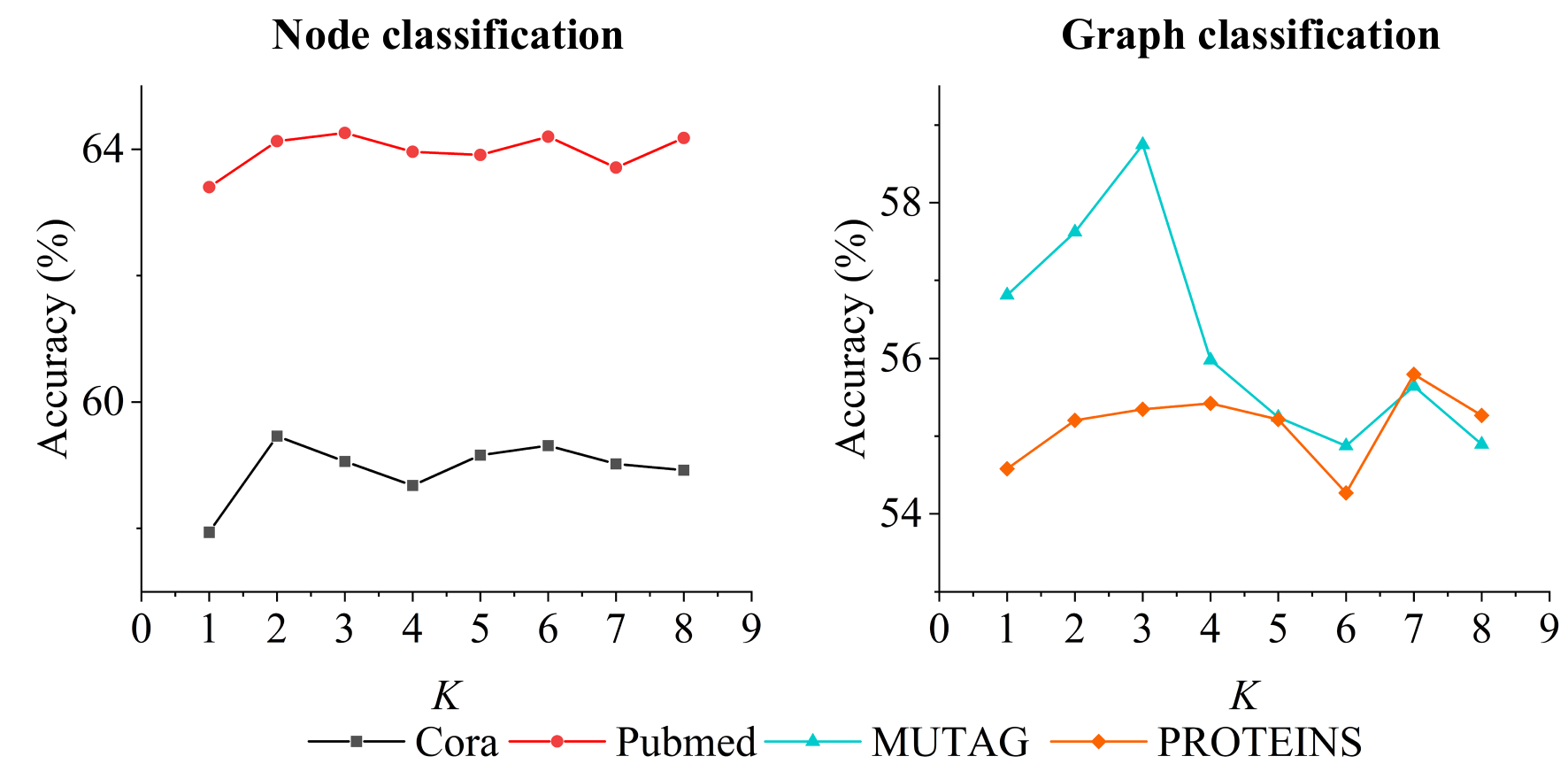}
\caption{Impact of number of inference steps $K$.}
\label{fig.think}
\end{figure}

\stitle{Number of inference steps.}
We further investigate the influence of the number of inference steps, \(K\), with the results summarized in Fig.~\ref{fig.think}. For node classification, we observe that \(K=2\) generally yields the best performance, whereas \(K=3\) is optimal for graph classification. This discrepancy may stem from the fact that graph classification relies more on global structural information, which may be better captured with more inference steps. 
Notably, for the \textsc{Proteins} dataset, \(K=7\) achieves the highest performance, likely due to its intrinsic complexity and variability, as discussed in Sect.~\ref{sec.exp.per}.

\subsection{Computational Efficiency Analysis}
To evaluate the efficiency of \model, we perform one-shot node classification on a large-scale dataset called CLUSTER \cite{dwivedi2023benchmarking}, which comprises 1,172,320 nodes and 43,038,630 edges. 
The accuracy evaluation is shown in Table~\ref{table.large-dataset}, whereas inference time comparisons with strong baselines are reported in Table~\ref{table.time}.

\begin{table}[tbp] 
    \centering
    \small
    \caption{Accuracy (\%) evaluation of node classification on the large-scale CLUSTER dataset.
    }
    \label{table.large-dataset}%
    \begin{tabular}{l|ccccc}
    \toprule
  Methods & \textsc{DGI} & \textsc{GraphCL} & \textsc{GraphPrompt} & \textsc{GPF+} &\model  \\\midrule
  Accuracy &21.27 &23.82 &22.18 &24.27 &\textbf{27.54}
    \\\bottomrule
    \end{tabular}
\end{table}

While \model\ maintains higher accuracy on the large-scale dataset, we observe additional inference time due to its multi-step inference on the test set. 
Such a limitation of CoT has also been observed in NLP. In our case, inference time increases roughly linearly with the number of steps, which is acceptable considering the significant accuracy gains.


\begin{table}[tbp] 
    \centering
    \small
    \caption{Inference time (milliseconds) on the test set of a node classification task.
    }
    \label{table.time}%
    \begin{tabular}{l|rrrr}
    \toprule
  {Methods}   & Cora & Pubmed &MUTAG &CLUSTER  \\\midrule
    \textsc{GraphPrompt} & 3.90 &5.62 & 5.85 &955.25\\
    \textsc{GPF+} & 5.25	& 5.92 & 4.79 &713.08\\
    \model & 7.95 & 13.95 & 11.44 &2340.00\\
    \bottomrule
        \end{tabular}
\end{table}

\section{Conclusions}
In this paper, we proposed \model, the first CoT prompting framework for text-free graphs. Specifically, we introduced the notion of step-by-step inference, where each inference step consists of three substages: prompt-based inference, thought construction, and thought-conditioned prompt learning. Concretely, we first feed the prompt-modified query graph into a pre-trained graph encoder, and then construct a thought by fusing the hidden embeddings from all layers of the pre-trained encoder to capture hierarchical structural knowledge.
Subsequently, conditioned on the thought, we generate a series of node-specific prompts to guide the next inference step. After multiple such inference steps, \model\ makes a prediction on the final ``answer.'' Lastly, we conducted extensive experiments on eight public datasets, demonstrating that \model\ significantly outperforms a range of state-of-the-art baselines.

\section*{Acknowledgments}
 This research / project is supported by the Ministry of Education, Singapore, under its Academic Research Fund Tier 2 (Proposal ID: T2EP20122-0041). Any opinions, findings and conclusions or recommendations expressed in this material are those of the author(s) and do not reflect the views of the Ministry of Education, Singapore. 

 \clearpage

\bibliographystyle{ACM-Reference-Format}
\bibliography{references}

\appendix
\section*{Appendices}
\renewcommand\thesubsection{\Alph{subsection}}
\renewcommand\thesubsubsection{\thesubsection.\arabic{subsection}}
\subsection{Further Descriptions of Datasets} \label{app.dataset}
We provide more comprehensive descriptions of the benchmark datasets\footnote{\url{https://github.com/shchur/gnn-benchmark/raw/master/data/npz/}}\footnote{\url{https://huggingface.co/datasets/graphs-datasets/MUTAG}}\footnote{\url{https://chrsmrrs.github.io/datasets/docs/datasets/}} used in our experiments.

\begin{itemize}[leftmargin=*]
    \item \emph{Cora} \cite{mccallum2000automating} is a citation network composed of 2,708 research papers in the field of computing, each classified into one of seven categories. The network consists of 5,429 citation links between papers. Each paper is represented by a binary word vector, where each entry indicates the presence or absence of a word from a predefined vocabulary of 1,433 unique terms.
    \item \emph{Citeseer} \cite{sen2008collective} includes 3,312 computer science publications, categorized into six distinct classes, separate from those in \textit{Cora}. The citation network contains 4,732 edges. Each document is encoded as a binary word vector that captures the presence or absence of words from a dictionary comprising 3,703 unique terms.
    \item \emph{PubMed} \cite{sen2008collective} is a citation network of 19,717 biomedical articles related to diabetes, divided into three categories. The network includes 44,338 citation edges. Unlike \textit{Cora} and \textit{Citeseer}, each document is represented by a TF/IDF-weighted word vector derived from a dictionary of 500 unique terms.
    \item \emph{Photo} \cite{shchur2018pitfalls} consists of 7,487 photography-related products, each assigned to one of eight categories. The co-purchase network contains 119,043 edges, where connections indicate products frequently bought together. Each product is described by a feature vector extracted from metadata and customer reviews, with category labels corresponding to product types.
    \item \textit{MUTAG} \cite{debnath1991structure} is a dataset of nitroaromatic compounds aimed at predicting their mutagenic effects on \textit{Salmonella typhimurium}. Each compound is modeled as a graph, where nodes represent atoms with categorical labels (encoded as one-hot vectors) based on atom types, and edges depict the chemical bonds connecting them. The dataset comprises 188 molecular graphs with 7 unique node types.
    
    \item \emph{BZR} \cite{nr} consists of 405 molecular graphs representing ligands that interact with benzodiazepine receptors. Each molecule is treated as an independent graph and is classified into one of two categories.
    
    \item \emph{COX2} \cite{nr} includes 467 molecular structures of cyclooxygenase-2 inhibitors. In this dataset, nodes correspond to atoms, while edges define chemical bonds—which may be single, double, triple, or aromatic. The molecules are divided into two distinct classes.
    
    \item \emph{PROTEINS} \cite{borgwardt2005protein} is a dataset of protein structure graphs that encode both biochemical and structural properties. In this dataset, nodes represent secondary structural elements, while edges capture connectivity based on spatial proximity or amino acid sequence adjacency. Each node falls into one of three categories, and graphs are classified into two broader groups.
\end{itemize}

\subsection{Further Descriptions of Baselines} \label{app.baselines}
We provide additional details about the baseline methods used in our experiments.

\vspace{1mm}
\noindent (1) Supervised GNNs.
\begin{itemize}[leftmargin=*]
    \item \textbf{GCN} \cite{kipf2016semi}: A graph neural network that aggregates node information using mean-pooling, thereby enabling nodes to capture structural information from their neighbors.
    \item \textbf{GAT} \cite{velivckovic2017graph}: Unlike GCN, GAT incorporates attention mechanisms to assign different weights to neighboring nodes, refining the aggregation process based on their relative importance.
\end{itemize}

\vspace{1mm}
\noindent(2) Graph Pre-training Models.
\begin{itemize}[leftmargin=*]
    \item \textbf{DGI} \cite{velivckovic2017graph}: A self-supervised pre-training method that maximizes mutual information between local node embeddings and the graph’s global representation, thereby enhancing structural awareness.
    \item \textbf{InfoGraph} \cite{sun2019infograph}: An extension of DGI designed for graph-level tasks, aligning node and graph representations by optimizing their similarity.
    \item \textbf{GraphCL} \cite{you2020graph}: A contrastive learning framework that leverages diverse graph augmentations to extract structural patterns, aiming to improve representation consistency across transformations.
\end{itemize}

\vspace{1mm}
\noindent(3) Standard Graph Prompting Models.
\begin{itemize}[leftmargin=*]
    \item \textbf{ProG} \cite{sun2023all}: Reformulates node- and edge-level tasks as graph-level problems by employing prompt graphs with task-specific structures to guide adaptation.
    \item \textbf{GPF} \cite{fang2024universal}: A universal prompt-tuning strategy for pre-trained graph models that transforms input graph features to mimic various prompting effects.
    \item \textbf{GPF+} \cite{fang2024universal}: An enhanced version of GPF that integrates an attention mechanism to dynamically refine prompt representations.
    \item \textbf{GraphPrompt} \cite{liu2023graphprompt}: Bridges pre-training and downstream tasks using subgraph similarity-based prompting, where a learnable prompt is optimized to incorporate task-relevant information for both node and graph classification.
\end{itemize}

\end{document}